\documentclass[sigconf]{acmart}
\AtBeginDocument{%
  }

\acmYear{2025} 
\setcopyright{cc}
\setcctype{by-nc-sa}
\acmConference[SIGIR-AP 2025]{Proceedings of the 2025 Annual International
ACM SIGIR Conference on Research and Development in Information Retrieval in
the Asia Pacific Region}{December 7--10, 2025}{Xi'an, China}
\acmBooktitle{Proceedings of the 2025 Annual International ACM SIGIR
Conference on Research and Development in Information Retrieval in the Asia
Pacific Region (SIGIR-AP 2025), December 7--10, 2025, Xi'an,
China}\acmDOI{10.1145/3767695.3769518}
\acmISBN{979-8-4007-2218-9/2025/12}

\usepackage{enumitem}
\usepackage{amsmath, amsthm}
\usepackage{balance}
\usepackage{amsfonts}
\usepackage{graphicx}
\usepackage{multirow}
\usepackage{multicol}
\usepackage{booktabs}
\usepackage{arydshln}
\usepackage{color}
\usepackage{makecell}
\usepackage{subcaption}
\usepackage{xcolor}[dvipsnames]
\usepackage{epigraph}
\usepackage{tcolorbox}
\usepackage{colortbl}
\definecolor{mygray}{gray}{.9}
\definecolor{mygreen}{rgb}{0,0.7,0}
\definecolor{myorange}{RGB}{255, 218, 185}
\definecolor{mycolor}{rgb}{0.8157, 0.251, 0.2196}

\begin{document}

\title{On the Diminishing Returns of Complex Robust RAG Training in the Era of Powerful LLMs}

\author{Hanxing Ding}
\authornote{Both authors contributed equally to this research.}
\email{dinghanxing18s@ict.ac.cn}
\affiliation{%
  \institution{State Key Laboratory of AI Safety, Institute of Computing Technology, Chinese Academy of Sciences}
  \city{Beijing}
  \country{China}
}

\author{Shuchang Tao}
\authornotemark[1]
\email{taoshuchang.tsc@alibaba-inc.com}
\affiliation{%
  \institution{Tongyi Lab, Alibaba Group}
  \city{Beijing}
  \country{China}
}

\author{Liang Pang}
\authornote{Corresponding author}
\email{pangliang@ict.ac.cn}
\affiliation{%
  \institution{State Key Laboratory of AI Safety, Institute of Computing Technology, Chinese Academy of Sciences}
  \city{Beijing}
  \country{China}
}

\author{Zihao Wei}
\affiliation{%
  \institution{Institute of Computing Technology}
  \city{Beijing}
  \country{China}
}

\author{Liwei Chen}
\affiliation{%
 \institution{Kuaishou Technology}
 \city{Beijing}
 \country{China}
}

\author{Kun Xu}
\affiliation{%
 \institution{Kuaishou Technology}
 \city{Beijing}
 \country{China}
}

\author{Huawei Shen}
\affiliation{%
  \institution{Institute of Computing Technology}
  \city{Beijing}
  \country{China}
}

\author{Xueqi Cheng}
\affiliation{%
  \institution{Institute of Computing Technology}
  \city{Beijing}
  \country{China}
}

\renewcommand{\shortauthors}{Ding et al.}

\begin{abstract}
  Retrieval-augmented generation (RAG) systems traditionally employ sophisticated training strategies to enhance robustness against retrieval noise. In this work, we investigate a critical question: does the benefit of these complex robust training methods diminish as language models become more powerful? Through systematic evaluation across multiple model scales and question-answering datasets, our analysis reveals a consistent trend: \emph{the marginal robustness benefit of sophisticated training strategies decreases substantially as model capacity increases.} While smaller models show significant performance improvements from complex document selection and adversarial objectives, more capable models achieve comparable or even superior performance with simpler training approaches. Further investigation demonstrates that stronger models naturally exhibit better confidence calibration, cross-dataset generalization capability, and more effective attention patterns, even under simple training regimes. These findings suggest that as foundation models evolve, the engineering effort invested in complex robust training may yield diminishing returns, indicating that simplified RAG pipelines could suffice for powerful models while maintaining competitive performance.
\end{abstract}

\ccsdesc[500]{Information systems~Question answering}

\keywords{Large Language Models, Retrieval-augmented Generation, Robust Training}


\maketitle

\section{Introduction}
Modern LLMs excel in various NLP tasks including text generation, knowledge reasoning, and question answering~\cite{DBLP:conf/lkm/YeLZHJ24,DBLP:conf/acl/TaoYDXC0GSD24,DBLP:journals/corr/abs-2303-18223}, yet they struggle with external knowledge integration, particularly in RAG systems where poor retrieval can lead to inaccurate responses~\cite{DBLP:conf/emnlp/XuQGW0ZX24,DBLP:conf/acl/Tan0YWCC24,DBLP:journals/corr/abs-2402-10612}. To address this challenge, researchers have proposed robust training methods focusing on high-quality document selection or adversarial loss regularization~\cite{DBLP:conf/iclr/YoranWRB24,DBLP:conf/acl/FangBNY0X24,DBLP:conf/icml/KruegerCJ0BZPC21,DBLP:conf/emnlp/ZhuYSYS24,DBLP:conf/acl/Jin00D24,DBLP:journals/corr/abs-2411-14572}. For instance, RetRobust~\cite{DBLP:conf/iclr/YoranWRB24} uses a mixture of relevant and irrelevant documents during training, while RAAT~\cite{DBLP:conf/acl/FangBNY0X24} and ATM~\cite{DBLP:conf/emnlp/ZhuYSYS24} incorporate adversarial regularization terms to maintain consistent performance across varied document contexts.

\begin{figure}[t]
    \centering
    \includegraphics[width=\columnwidth]{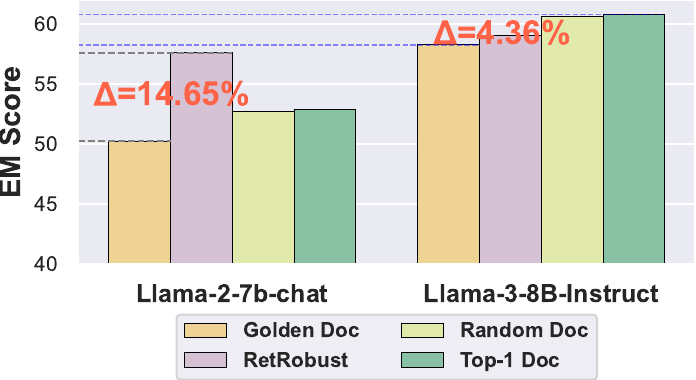}
    \caption{Performance comparison of different robust training strategies on the TriviaQA dataset. As model capabilities increase, the marginal robustness benefit ($\Delta$), i.e., performance gain between the best and worst strategies among the four training methods decreases from 14.65\% to 4.36\%.}
    \label{fig:intro}
    \vspace{-0.3cm}
\end{figure}

With the rapid advancement of LLM capabilities, a critical research question arises: \textit{how does the effectiveness of sophisticated robust training strategies change as model capacity grows?} 

To investigate this question empirically, we conduct a preliminary analysis comparing four distinct training strategies on the TriviaQA dataset, as illustrated in Figure~\ref{fig:intro}. Our results reveal a striking pattern: the performance gain $\Delta$ between the best and worst strategies decreases from 14.65\% to 4.36\% as model capacity scales up. This pronounced reduction in $\Delta$ suggests a law of diminishing returns, wherein the marginal benefit of complex training methods becomes increasingly negligible as model capacity expands.

In this paper, building upon this observation, we conduct a systematic investigation to validate and characterize the diminishing returns of robust RAG training as model capacity increases. We propose an evaluation framework that quantifies the \emph{marginal robustness benefit} ($\Delta$)—the performance gain introduced by sophisticated document selection and adversarial training methods. Our comprehensive experiments confirm and strengthen the initial finding: \emph{as model capability increases, the marginal robustness benefit of complex training diminishes sharply.} While smaller models significantly benefit from curated documents and adversarial losses, stronger models achieve comparable or even superior performance with simpler setups. Strikingly, training high-capacity models with randomly selected documents often matches or exceeds the performance of advanced robust training methods, challenging fundamental assumptions about robust training strategies.

To uncover the mechanisms behind this phenomenon, we perform a comprehensive analysis of model behavior. Our findings show that stronger models possess inherent capabilities that reduce reliance on complex training. They exhibit superior confidence calibration, accurately distinguishing correct from incorrect outputs without additional supervision. Even when trained with random documents, these models generalize well across datasets. Attention visualizations further reveal that high-capacity models naturally learn effective attention patterns under simple training, explaining the diminishing gains from sophisticated methods.


These findings reveal a shift in how robust RAG systems should be designed: as model capacity increases, simpler training and retrieval strategies become not only sufficient but preferable. This shift opens up new opportunities for scalable RAG applications, and prompts a rethinking of training priorities in the context of model scaling.

The main contributions of this work are:
\begin{itemize}
    \item \textbf{Finding:} We demonstrate that the marginal benifit from sophisticated robust training diminish significantly as model capacity increases.
    \item \textbf{Rationale:} Through detailed analysis, we attribute this finding to intrinsic properties of stronger models-including better confidence calibration, cross-task generalization, and effective attention patterns—even under simple robust training.
\end{itemize}

\section{Preliminaries}
\subsection{Related Work}
\label{app:related_works}
\subsubsection{Retrieval-Augmented Generation}
\label{app:related_rag}
Retrieval-Augmented Generation has been developed to enhance the reasoning and generation capabilities of LLMs by integrating external information. However, dense retrievers are not always perfect and can sometimes recall irrelevant or erroneous documents, which can degrade performance~\cite{DBLP:conf/emnlp/JiangXGSLDYCN23,DBLP:journals/corr/abs-2402-10612}. To mitigate this issue, adaptive retrieval techniques and robust RAG training methods have been proposed. Adaptive retrieval techniques retrieve external information only when LLMs encounter unknowns, reducing the risk of integrating misleading or incorrect information~\cite{DBLP:conf/emnlp/JiangXGSLDYCN23,DBLP:conf/acl/MallenAZDKH23,DBLP:journals/corr/abs-2310-11511}. Additionally, robust training focuses on enhancing LLMs' response to various retrieval noises, aiming to efficiently obtain adversarial examples that improve model robustness while reducing training overhead~\cite{DBLP:conf/acl/FangBNY0X24,DBLP:conf/emnlp/ZhuYSYS24,DBLP:journals/corr/abs-2405-15556}.

\subsubsection{Robust Training for RAG Systems}
\label{app:related_robusttraining}
The performance of LLMs in RAG systems can be compromised by irrelevant context~\cite{DBLP:conf/sigir/CuconasuTSFCMTS24,DBLP:journals/corr/abs-2408-13533}. Recently, researchers have focused on this issue and proposed effective adversarial training methods to enhance the robustness of LLMs against noisy documents. RetRobust~\cite{DBLP:conf/iclr/YoranWRB24} suggests using a mix of relevant and irrelevant documents to train LLMs to withstand noisy contexts. RAAT~\cite{DBLP:conf/acl/FangBNY0X24} and ATM~\cite{DBLP:conf/emnlp/ZhuYSYS24} employ adversarial training to dynamically adjust the model’s training process, aiming to maintain robustness against noise and generate accurate answers. RobustRAG~\cite{DBLP:journals/corr/abs-2405-15556} proposes to enhances RAG model robustness against retrieval corruption attacks through an isolate-then-aggregate strategy to achieve certifiable robustness. Additionally, another work focus on identifying key documents from noisy contexts to facilitate effective question answering through knowledge rewriting, refinement, or filtering~\cite{DBLP:conf/acl/0025FDGY0CCC024,DBLP:journals/corr/abs-2406-08116,DBLP:conf/acl/Jin00D24,DBLP:journals/corr/abs-2411-14572}. However, these robust training approaches are primarily applied to small or weak LMs with fewer than 7 billion parameters. Thus, there’s an urgent need to explore whether complex robust training is still necessary to improve the robustness and generalization of bigger or stronger models when dealing with noisy contexts.

\subsection{Robustness Training for RAG}
\label{sec:robust_training}
\subsubsection{Retrieval-augmented Adaptive Adversarial Training}
\label{app:methods_adaptive}
To improve the robustness of retrieval-augmented language models (RALMs) against retrieval noise, RAAT~\cite{DBLP:conf/acl/FangBNY0X24} incorporates an adversarial loss into the standard supervised fine-tuning (SFT). The model processes one golden context and three adversarial samples at each iteration, optimizing for the most challenging perturbation. The adversarial objective follows a min-max strategy:
\begin{equation}
\textstyle
\setlength\abovedisplayskip{0.2cm}
\setlength\belowdisplayskip{0.2cm}
\begin{aligned}
\mathcal{L}_{\text{max}} = \max_{da \in DA} \mathcal{L}(\theta, da(q), a),
\end{aligned}
\end{equation}
where $\mathcal{L}$ represents the generation loss function to noisy retrievals, and $q^{\prime}=da(q)$ represents the augmented noise context of $q$. A regularization term controls excessive sensitivity to noise:
\begin{equation}
\textstyle
\setlength\abovedisplayskip{0.2cm}
\setlength\belowdisplayskip{0.2cm}
\begin{aligned}
\mathcal{L}_{\text{ada}} = \mathcal{L}_{\text{max}} + w_{\text{reg}} \cdot |\mathcal{L}_{\text{max}} - \mathcal{L}_{\text{min}}|_2^2,
\end{aligned}
\end{equation}

The adversarial loss $\mathcal{L}_{\text{ada}}$ combines both terms, where the regularization term helps stabilize training by mitigating excessive sensitivity to retrieval noise. The regularization term, calculated as the square of the difference between $\mathcal{L}_{\text{max}}$ and $\mathcal{L}_{\text{min}}$, encourages a more balanced optimization, preventing the model from overreacting to the most challenging perturbations.

\subsubsection{Multiagent Iterative Tuning Optimization}
\label{app:methods_multi}
ATM~\cite{DBLP:conf/emnlp/ZhuYSYS24} steers the generator to have a robust perspective of useful documents for question answering with the help of an auxiliary attacker agent. The generator is trained to maximize answer correctness while minimizing its sensitivity to adversarial perturbations. It receives the user query along with a document list, which may contain both relevant and fabricated information introduced by the attacker. Its objective is to generate accurate responses as long as sufficient truthful information is present while reducing the impact of misleading or fabricated content. To achieve this, the generator learns to identify and leverage relevant documents while ignoring noisy ones, regardless of whether they originate from the original retrieval $\mathcal{D}$ or adversarial perturbations $\mathcal{D}^{\prime}$. This can be formalized as maximizing the objective:
\begin{equation}
\setlength\abovedisplayskip{0.2cm}
\setlength\belowdisplayskip{0.2cm}
\begin{aligned}
G(a\mid q,\mathcal{D}^{\prime}) - \mathrm{dist}\left[G(a\mid q, \mathcal{D}), G(a\mid q,\mathcal{D}^{\prime})\right],
\end{aligned}
\end{equation}
where ${G}(\cdot)$ represents the language model probability of generating an answer, and $\mathrm{dist}\left[\cdot\right]$ measures the divergence between outputs under different document conditions. The generator and the attacker are tuned adversarially for several iterations. After rounds of multi-agent iterative tuning, the generator can eventually better discriminate useful documents amongst fabrications.

\subsubsection{Invariant Risk Minimization}
\label{app:methods_irm}
Invariant Risk Minimization (IRM) aims to learn representations that remain stable across different environments, improving generalization under distribution shifts. To enhance the robustness of retrieval-augmented generation (RAG) across varying retrieval conditions, the V-REx objective~\cite{DBLP:conf/icml/KruegerCJ0BZPC21} can also be adapted to enforce risk invariance across different retrieval environments. Given retrieval environments $\mathcal{E} = \{1, \dots, m\}$, where each $e \in \mathcal{E}$ corresponds to a specific retrieval scenario (e.g., golden documents, top-$k$ retrieved, noisy retrieval), we define the empirical risk $\mathcal{R}_e(\theta)$ of the model parameterized by $\theta$. The training objective is formulated as:
\begin{equation}
\setlength\abovedisplayskip{0.2cm}
\setlength\belowdisplayskip{0.2cm}
\begin{aligned}
\mathcal{R}_\textrm{V-REx-RAG}(\theta) = &\beta \; \mathrm{Var}(\{\mathcal{R}_1(\theta), ..., \mathcal{R}_m(\theta)\})\\ & + \sum^m_{e=1} \mathcal{R}_e(\theta),
\end{aligned}
\end{equation}
where $\beta \geq 0$ controls the trade-off between minimizing average risk and enforcing risk invariance across retrieval environments. A higher $\beta$ reduces performance discrepancies caused by retrieval variations, improving generalization under both high-quality and noisy retrieval conditions.

\section{Evaluating Marginal Benefit of Robust RAG Methods}

In this section, we establish a systematic evaluation framework to address a fundamental question in RAG: whether sophisticated robust training methods continue to yield substantial performance improvements as model capabilities increase. We first formalize our evaluation metric to quantify the marginal benefit provided by different robust RAG strategies, followed by a detailed categorization of current state-of-the-art robust RAG methods. We also provide an experimental setup that ensures fair comparison across model scales and dataset types.

\subsection{Metric: Marginal Robustness Benefit}
To measure the relative utility of sophisticated robust training methods, we introduce the Marginal Robustness Benefit, denoted as $\Delta$, which captures the relative performance gain between the best and worst robust RAG strategies:
\begin{equation}
\Delta = \frac{S_{\text{best}} - S_{\text{worst}}}{S_{\text{worst}}}
\end{equation}
\noindent where $S_{\text{best}}$ and $S_{\text{worst}}$ denote the performance scores (EM or F1) of the most and least effective training strategies, respectively.

By computing $\Delta$ across models of varying capacities, we quantify how the relative effectiveness of robust training evolves with model scaling. This metric helps assess whether the complexity introduced by advanced training strategies remains necessary as foundation models grow more capable.

\subsection{Taxonomy of Robust RAG Methods}
We categorize current robust RAG training approaches into two dimensions:
\subsubsection{Document Selection Strategy}
We first analyze robust RAG training methods based on document selection strategies. These approaches use different criteria to select documents for supervised fine-tuning, thereby improving model adaptability to retrieval variations:
\begin{itemize}[leftmargin=0.5cm, itemindent=0cm, itemsep=0pt]
\item \textbf{Golden Document}: Selects the most relevant document containing the correct answer. If none exist in the top-20 documents, use the top-1 document for consistency.
\item \textbf{Top-1 Document}: Uses the highest-scoring retrieved document, which may not contain the correct answer, reflecting real-world retrieval challenges.
\item \textbf{RALM}: Fine-tunes the model by prepending golden retrieval text. Queries without golden documents in the top-20 docs are excluded.
\item \textbf{RetRobust}: Following \citet{DBLP:conf/iclr/YoranWRB24}, this method enhances robustness by randomly selecting top-ranked, low-ranked, or random passages for training.
\end{itemize}

To assess robustness against irrelevant information, we also include two adversarial scenarios:
\begin{itemize}[leftmargin=0.5cm, itemindent=0cm, itemsep=0pt]
\item \textbf{Random Document}: Randomly selects a document from retrieved results.
\item \textbf{Irrelevant Document}: Uses a passage from another query's results, ensuring complete irrelevance.
\end{itemize}

\subsubsection{Adversarial Loss Design}
Next, we analyze robust RAG methods based on adversarial loss designs. These approaches enhance model resilience against retrieval noise by combining standard SFT with robustness-enforcing regularization. Models train across multiple retrieval environments (golden documents, top-k retrievals, or adversarially perturbed documents) using a dual objective: standard training loss plus regularization to reduce sensitivity to retrieval variations.

\begin{table}[t]
\centering
\resizebox{0.85\linewidth}{!}{
\begin{tabular}{lccc}
\toprule
\textbf{Dataset} & \textbf{Type} & \textbf{\# Train} & \textbf{\# Dev} \\
\midrule
NQ & single-hop & 79,168 & 8,757 \\
WebQuestions & single-hop & 2,474 & 278 \\
TriviaQA & multi-hop & 78,785 & 8,837 \\
HotpotQA & multi-hop & 90,447 & 7,405 \\
\bottomrule
\end{tabular}}
\caption{Statistics of different datasets}
\label{tab:dataset_stats}
\vspace{-0.7cm}
\end{table}

We evaluate two adversarial loss strategies:
\begin{itemize}[leftmargin=0.5cm, itemindent=0cm, itemsep=0pt]
\item \textbf{RAAT}: Reduces performance gap between best and worst retrieval cases, ensuring stability under challenging conditions.
\item \textbf{IRM}: Minimizes performance variance across retrieval environments, mitigating sensitivity to distribution shifts for reliable cross-scenario performance.
\end{itemize}

\subsection{Experimental Setups}

\begin{table*}[t]
\centering
\resizebox{0.95\linewidth}{!}{
\begin{tabular}{clcccccccccc}
\toprule[1.5pt]
\multirow{2}{*}{\textbf{Model}} & \multirow{2}{*}{\textbf{RAG Methods}} & \multicolumn{2}{c}{\textbf{HotpotQA}} & \multicolumn{2}{c}{\textbf{NQ}} & \multicolumn{2}{c}{\textbf{WebQuestions}} & \multicolumn{2}{c}{\textbf{TriviaQA}} & \multicolumn{2}{c}{\textbf{AVERAGE}}\\
\cline{3-4}\cline{5-6}\cline{7-8}\cline{9-10}\cline{11-12}
 & &  \textbf{EM} & \textbf{F1} & \textbf{EM} & \textbf{F1} & \textbf{EM} & \textbf{F1} & \textbf{EM} & \textbf{F1} & \textbf{EM} & \textbf{F1} \\
\midrule[1pt]
\multirow{10}{*}{\rotatebox{90}{\textit{Llama-2-7b-chat-hf}}} & Base Model &  3.30  &  12.34 &  1.21 &  10.61  &  0.00  &  13.08  &  4.32  &  20.27 & 2.21 & 14.08  \\
\cdashline{2-12}
 & RALM &  26.21   & 36.42  &  32.17 &  42.68 &  33.81  &  45.85  &  50.28  &  60.17 & 35.62 & 46.28  \\
 & RetRobust &  31.29  &  43.65  &  37.71  & 49.49  &  36.33  &  47.98  &  57.61  & 67.52 & 40.74 & 52.16  \\
 & Top-1 Doc & 31.76 & 43.95 & 40.20 & 51.89  &  41.73  &  52.76  &  52.93  & 65.41 & 41.66 & 53.50  \\
 & Golden Doc & 30.67 & 42.78 & 36.50 & 47.77  &  39.93  &  52.11  &  50.25  & 63.28 & 39.34 & 51.49  \\
 & Random Doc & 30.94 & 43.11 & 38.16 & 49.78  &  42.45  &  53.97  &  52.72 &  65.52 & 41.07 & 53.10  \\
 & Irrelevant Doc & 31.01 & 42.98 & 37.08 & 48.93  &  39.21   &  50.79  &  51.97   &  64.70 & 39.82 & 51.85 \\
 & RAAT & 31.32 & 43.24 & 42.91 & 53.19  &  36.69  &  48.82 &  51.65 &  58.71 & 40.64 & 50.99  \\
 & IRM & 34.38 & 47.11 & 40.96 & 53.07  &  53.96  &  61.62 &  57.58 &  69.08 & 46.72 & 57.72  \\
 \cmidrule{2-12}
& \textcolor{mycolor}{$\Delta$ (Worst $\rightarrow$ Best)}  &  \textcolor{mycolor}{31.17\%}  & \textcolor{mycolor}{29.35\%}  &  \textcolor{mycolor}{33.39\%}  &  \textcolor{mycolor}{24.63\%}  &  \textcolor{mycolor}{59.60\%}   &  \textcolor{mycolor}{34.39\%}  &  \textcolor{mycolor}{14.65\%}  &  \textcolor{mycolor}{17.66\%} & \textcolor{mycolor}{31.16\%} & \textcolor{mycolor}{24.72\%}  \\
\midrule
\multirow{10}{*}{\rotatebox{90}{\textit{Llama-3-8B-Instruct}}} & Base Model & 23.31 & 32.60 & 30.04 & 41.59 & 26.98 & 43.25 & 58.80 & 66.45 & 34.78 & 45.97 \\
\cdashline{2-12}
 & RALM &  30.73 &  41.25  &  35.19  &  46.10  &   47.84  &  56.98   &  54.75 &  63.27  & 41.36 & 51.11 \\
 & RetRobust &  36.88 &  49.43  &  43.28  &  55.04  &  52.88  &  62.10  &  59.06  & 67.77 & 48.03 & 58.59  \\
 & Top-1 Doc & 37.09 & 49.67 & 44.38 & 56.20  &  54.68  &  62.26 &  60.80  & 68.31 & 49.24 & 59.11  \\
 & Golden Doc & 35.68 & 48.97 & 41.35 & 53.13  &  48.92  &  58.41  &  58.26 & 66.99 &  46.05 & 56.88 \\
 & Random Doc & 36.73 & 49.53 & 43.37 & 55.43  &  53.24  &   62.55 &  60.62  & 68.64 & 48.49 & 59.04  \\
 & Irrelevant Doc & 35.53 & 48.27 & 42.45 & 54.41  &  46.76  &  57.67  &  58.97  & 66.57 & 45.93 & 56.73 \\
 & RAAT & 32.79 & 44.26 & 42.34 & 53.31  &  48.28  &  58.17 &  54.41 &  62.45 & 44.46 & 54.55  \\
 & IRM & 35.63 & 48.72 & 41.13 & 53.14  &  53.96  &  61.64  &  57.15 &  69.13 & 46.97 & 58.16  \\
\cmidrule{2-12}
& \textcolor{mycolor}{$\Delta$ (Worst $\rightarrow$ Best)}  &  \textcolor{mycolor}{20.70\%}  & \textcolor{mycolor}{20.41\%}  &  \textcolor{mycolor}{26.12\%}  &  \textcolor{mycolor}{21.91\%}  &  \textcolor{mycolor}{16.94\%}   &  \textcolor{mycolor}{9.78\%}  &  \textcolor{mycolor}{11.74\%}  &  \textcolor{mycolor}{10.70\%}  & \textcolor{mycolor}{16.88\%} & \textcolor{mycolor}{13.89\%} \\
\bottomrule[1.5pt]
\end{tabular}}
\caption{Performance comparison of Llama-2-7b-chat-hf and Llama-3-8B-Instruct across various robust RAG methods on four datasets (HotpotQA, NQ, WebQuestions, and TriviaQA). The row \textcolor{mycolor}{$\Delta$ (Worst $\rightarrow$ Best)} indicates the marginal robustness benefit achieved by the best method compared to the worst method.}
\label{tab:main_result_llama}
\vspace{-0.5cm}
\end{table*}
\begin{table*}[t]
\centering
\resizebox{0.95\linewidth}{!}{
\begin{tabular}{llcccccccccc}
\toprule[1.5pt]
\multirow{2}{*}{\textbf{Model}} & \multirow{2}{*}{\textbf{RAG Methods}} & \multicolumn{2}{c}{\textbf{HotpotQA}} & \multicolumn{2}{c}{\textbf{NQ}} & \multicolumn{2}{c}{\textbf{WebQuestions}} & \multicolumn{2}{c}{\textbf{TriviaQA}}  & \multicolumn{2}{c}{\textbf{AVERAGE}} \\
\cline{3-4}\cline{5-6}\cline{7-8}\cline{9-10}\cline{11-12}
&  &  \textbf{EM} & \textbf{F1} & \textbf{EM} & \textbf{F1} & \textbf{EM} & \textbf{F1} & \textbf{EM} & \textbf{F1} & \textbf{EM} & \textbf{F1}  \\
\midrule[1pt]
\multirow{9}{*}{\rotatebox{90}{\textit{Qwen1.5-7B-Chat}}} & Base Model & 24.15 & 34.06 & 24.31 & 35.27 & 23.02 & 38.65 & 48.43 & 57.65 & 29.98 & 41.41 \\
\cdashline{2-12}
& RALM & 24.44 & 34.23 & 29.83 & 40.22 & 44.24 & 53.46 & 46.40 & 55.96 & 36.23 & 45.97 \\
& RetRobust & 29.05 & 40.99 & 34.82 & 45.92 & 47.12 & 55.39 & 49.16 & 59.01 & 40.04 & 50.33 \\
& Top-1 Doc & 29.68 & 41.57 & 34.16 & 45.44 & 43.88 & 52.71 & 49.61 & 59.03 & 39.33 & 49.69 \\
& Golden Doc & 28.66 & 40.55 & 32.93 & 43.99 & 43.88 & 52.74 & 49.25 & 58.87 & 38.68 & 49.04 \\
& Random Doc & 28.83 & 40.66 & 33.96 & 45.09 & 46.04 & 53.21 & 48.75 & 58.54 & 39.40 & 49.38 \\
& Irrelevant Doc & 28.12 & 39.87 & 32.74 & 43.94 & 43.88 & 52.67 & 48.33 & 58.18 & 38.27 & 48.67 \\
& IRM & 26.95 & 38.67 & 30.59 & 41.83 & 47.84 & 55.68 & 45.52 & 56.24 & 37.73 & 48.11 \\ 
\cmidrule{2-12}
& \textcolor{mycolor}{$\Delta$ (Worst $\rightarrow$ Best)} &  \textcolor{mycolor}{21.44\%}  & \textcolor{mycolor}{21.44\%}  &  \textcolor{mycolor}{16.73\%}  &  \textcolor{mycolor}{14.17\%}  &  \textcolor{mycolor}{9.02\%}   &  \textcolor{mycolor}{5.71\%}  &  \textcolor{mycolor}{8.99\%}  &  \textcolor{mycolor}{5.49\%} & \textcolor{mycolor}{10.52\%} & \textcolor{mycolor}{9.48\%} \\
\midrule
\multirow{9}{*}{\rotatebox{90}{\textit{Qwen2.5-7B-Instruct}}} & Base Model & 25.10 & 35.02 & 26.97 & 38.15 & 25.90 & 42.46 & 53.86 & 62.73 & 32.96 & 44.59 \\
\cdashline{2-12}
& RALM & 26.37 & 36.30 & 31.43 & 42.00 & 41.73 & 51.98 & 53.92 & 62.74 & 38.36 & 48.26 \\
& RetRobust & 30.47 & 42.26 & 34.60 & 46.00 & 45.68 & 53.80 & 54.43 & 63.63 & 41.30 & 51.42 \\
& Top-1 Doc & 30.24 & 42.61 & 34.72 & 46.20 & 46.04 & 54.13 & 53.65 & 63.01 & 41.16 & 51.49 \\
& Golden Doc & 30.28 & 41.84 & 33.77 & 45.11 & 44.60 & 53.18 & 54.35 & 63.55 & 40.75 & 50.92 \\
& Random Doc & 30.25 & 42.22 & 34.08 & 45.47 & 44.24 & 53.30 & 54.29 & 63.54 & 40.72 & 51.13 \\
& Irrelevant Doc & 29.76 & 41.67 & 33.64 & 44.87 & 42.09 & 52.72 & 53.38 & 63.01 & 39.72 & 50.57 \\
& IRM & 29.58 & 41.73 & 32.79 & 44.17 & 46.40 & 54.80 & 53.45 & 64.26 & 40.56 & 51.24 \\
\cmidrule{2-12}
& \textcolor{mycolor}{$\Delta$ (Worst $\rightarrow$ Best)} &  \textcolor{mycolor}{15.55\%}  & \textcolor{mycolor}{17.38\%}  &  \textcolor{mycolor}{10.47\%}  &  \textcolor{mycolor}{10.00\%}  &  \textcolor{mycolor}{11.19\%}   &  \textcolor{mycolor}{5.43\%}  &  \textcolor{mycolor}{1.97\%}  &  \textcolor{mycolor}{2.42\%} & \textcolor{mycolor}{7.64\%}  &  \textcolor{mycolor}{6.70\%} \\
\bottomrule[1.5pt]
\end{tabular}}
\caption{Performance comparison of different LLMs (Qwen1.5-7B-Chat and Qwen2.5-7B-Instruct) across different robust RAG scenarios on four datasets (HotpotQA, NQ, WebQuestions, and TriviaQA).}
\label{tab:main_result_qwen}
\vspace{-0.6cm}
\end{table*}

\paragraph{Datasets} For our experiment, we evaluate on four widely-used question answering datasets: (1) single-hop QA, including NaturalQuestions (NQ)~\cite{DBLP:journals/tacl/KwiatkowskiPRCP19} and WebQuestions~\cite{DBLP:conf/emnlp/BerantCFL13}; and (2) multi-hop QA, including TriviaQA~\cite{DBLP:conf/acl/JoshiCWZ17} and HotpotQA~\cite{DBLP:conf/emnlp/Yang0ZBCSM18}. All experimental results are evaluated on their dev splits using the Exact Match (EM) and F1 metrics. Detailed statistics of these datasets are listed in Table~\ref{tab:dataset_stats}.

\paragraph{Implementation Details}  
We implement a standard RAG framework with retrieval and generation phases. For retrieval, we use Contriever~\cite{DBLP:journals/tmlr/IzacardCHRBJG22}, a BERT-based dense retriever to retrieve top-20 documents per query as a candidate pool. We sample different document subsets from the candidate pool to create varying retrieval environments for each methods. For question-answering, the top-5 retrieved documents are concatenated with the query as input for answer generation. For model training, we utilize the LLaMA-Factory library to facilitate efficient LLM finetuning. We train our models with a learning rate of 1e-6 and set the maximum number of epochs to 3. To optimize the training process, we employ bfloat16 precision and DeepSpeed ZeRO-3 for distributed training. The gradient accumulation steps are set to 2 with a batch size of 8. For inference, we leverage vLLM to ensure efficient model serving. We maintain the default decoding parameters for each model during inference, including the default temperature and top-p sampling probabilities, to ensure fair comparison across different model variants. All training and inference procedures are conducted on 8 $\times$ NVIDIA A100 80GB GPUs.

\section{Diminishing Benefits of Complex Methods With Model Capacity}
To investigate whether sophisticated document selection strategies and adversarial loss designs are still essential for robust RAG performance as LLMs continue to evolve, we conduct comprehensive experiments across multiple LMs and datasets.

\subsection{Marginal Benefits of Sophisticated Document Selection}

We conduct experiments to analyze the effectiveness of complex document selection strategies under Llama model families (Llama-2-7b-chat-hf and Llama-3-8B-Instruct) in Table~\ref{tab:main_result_llama} and Qwen model families (Qwen1.5-7B-Chat and Qwen2.5-7B-Instruct) in Table~\ref{tab:main_result_qwen}.

\paragraph{Training with sophisticated documents enhances LM robustness for weak models}
The experimental results provide compelling evidence that robust training significantly improves model resilience when processing noisy documents. While base models exhibit substantial performance degradation when encountering noisy documents during inference, models that undergo robust training maintain consistent and superior QA performance across various document selection strategies. A notable example is the Llama-2-7b-chat-hf model's performance (Table~\ref{tab:main_result_llama}) on the HotpotQA dataset,  where training with golden documents improves the EM score from 3.3 (Base Model) to 30.67 (Golden Doc), indicating increased resilience to document noise. This pattern of improvement is consistently observed across both Llama (Table~\ref{tab:main_result_llama}) and Qwen (Table~\ref{tab:main_result_qwen}) model families, strongly indicating that robust training effectively mitigates the base models' inherent vulnerability to noisy documents.

\paragraph{Training with random documents shows surprising effectiveness}
We also notice that training with randomly selected documents exhibits remarkable effectiveness across all experimental configurations. Quantitative analysis shows that with Llama-2-7b-chat-hf, this approach achieves superior performance on WebQuestions compared to more sophisticated strategies. Similar observations emerge from experiments with Qwen1.5-7B-Chat, where random document selection achieves 46.04 EM on WebQuestions, approaching the optimal performance of 47.12 EM achieved by RetRobust. The consistency of these results across distinct model architectures suggests that the efficacy of random document selection represents an inherent characteristic of contemporary RAG systems.

\paragraph{Diminishing returns of sophisticated document selections as models evolve}  
Experimental results indicate that the marginal robust benefit from sophisticated document selection strategies diminish as models evolve. For instance, comparing Llama-2 with Llama-3, the improvement in performance due to advanced document selection strategies decreases significantly, with the $\Delta$ (Worst $\rightarrow$ Best) metric for WebQuestions dropping from 59.60\% to 16.94\% EM. A similar trend is observed when comparing Qwen1.5 to Qwen2.5, where the performance improvement from sophisticated document selection also shows a noticeable reduction. These results suggest that as models become more advanced, their ability to process and utilize information improves independently of complex document selection strategies, leading to diminished returns from such methods.

\subsection{Marginal Benefits of Complex Loss Design}

To investigate whether the design of complex adversarial loss functions contributes to model performance, in Table~\ref{tab:main_result_llama} and~\ref{tab:main_result_qwen}, we also analyze the robustness of various adversarial loss designs.

\paragraph{Adversarial loss significantly enhances performance for weaker models}  
For the weaker model (Llama-2-7b-chat-hf), incorporating adversarial loss functions such as RAAT and IRM leads to a substantial improvement in performance compared to the base model or alternative document selection strategies. Specifically, while the base model achieves an average EM / F1 of only 2.21 / 14.08, applying adversarial loss functions boosts the performance to 40.64 / 50.99 for RAAT and 46.72 / 57.72 for IRM. This highlights the effectiveness of adversarial loss in improving model robustness to noisy documents, significantly enhancing both robustness and downstream inference performance. Notably, in some cases, RAAT and IRM outperform traditional document selection strategies (e.g., top-1 Doc and golden doc), demonstrating their value in scenarios where the model needs stronger guidance to handle noisy retrievals.

\begin{figure*}[t]
    \centering
    \includegraphics[width=0.98\columnwidth]{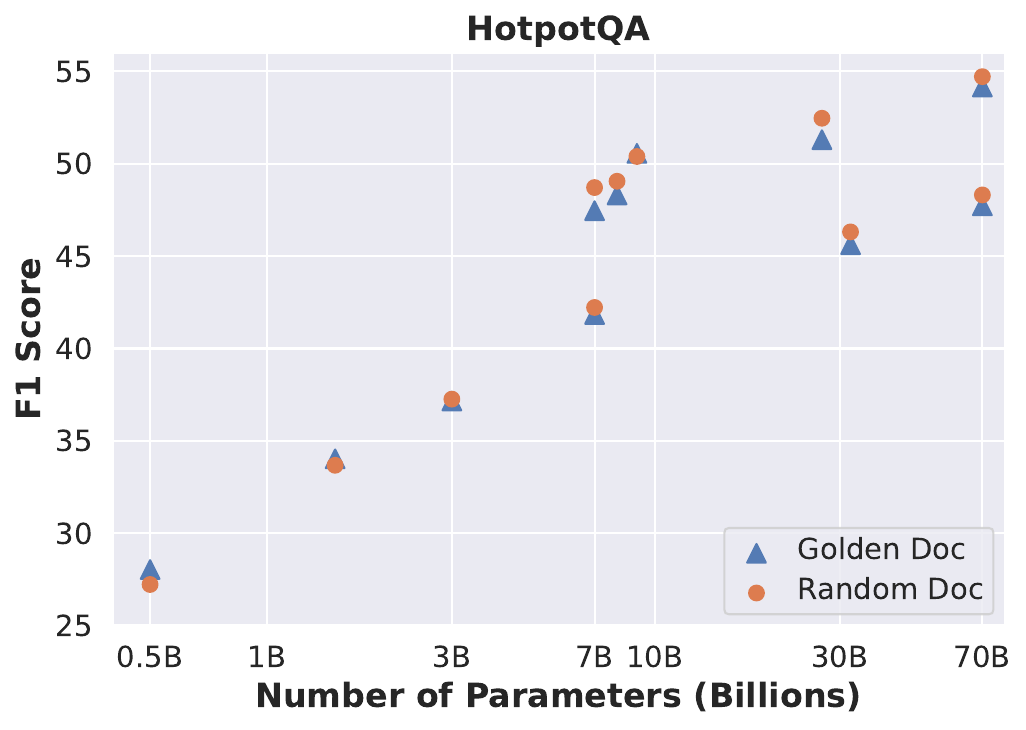}
    \includegraphics[width=0.98\columnwidth]{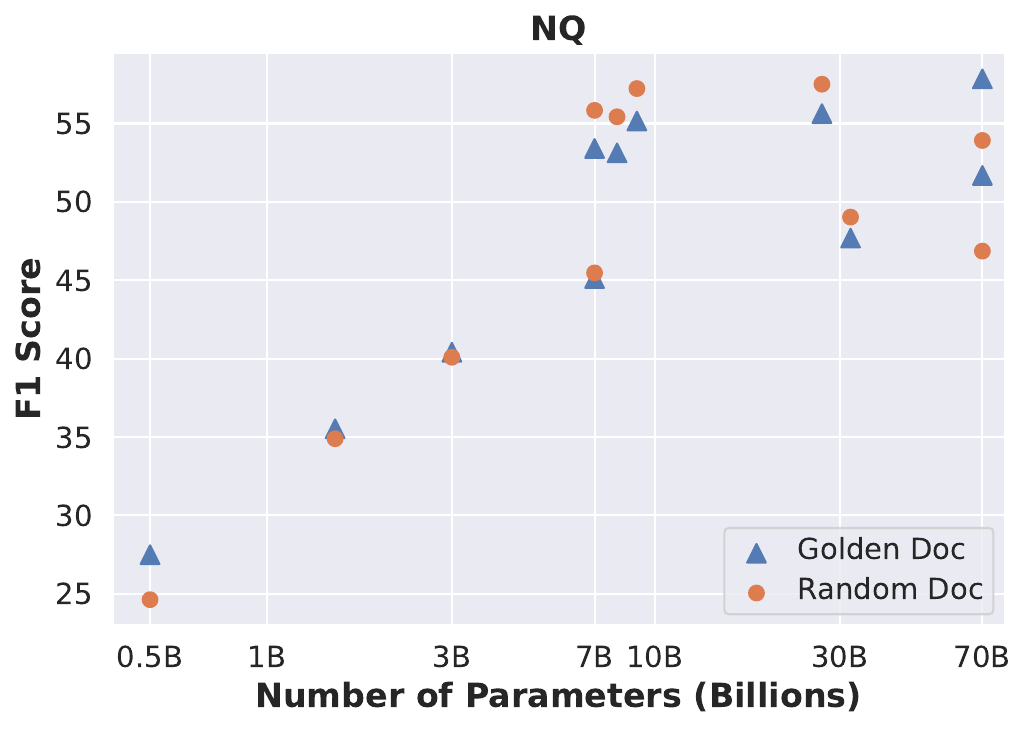}
    \caption{Comparison of F1 Scores on the HotpotQA (left) and NQ (right) dataset for training with Random Doc and Golden Doc across models with varying parameter sizes from 0.5B to 70B.}
    \label{fig:parameter_scale}
    \vspace{-0.3cm}
\end{figure*}

\paragraph{Adversarial loss exhibits diminishing returns for stronger models}  
For the stronger models, such as Llama-3-8B-Instruct and Qwen2.5-7B-Instruct, the benefits of adversarial loss functions are less pronounced. The basic Llama-3-8B-Instruct already achieves an average EM / F1 of 34.78 / 45.97, and the introduction of adversarial losses (RAAT and IRM) results in only modest improvements, with average EM / F1 scores of 44.46 / 54.55 and 46.97 / 58.16, respectively. This indicates that the models’ inherent robustness reduces the impact of adversarial losses, and in some cases, may even hinder performance. We hypothesize that when the model’s internal robustness is already well-developed, additional constraints from adversarial losses may interfere with its ability to optimize on clean and relevant inputs.

\paragraph{Conclusion: the necessity of adversarial loss.}
Based on these findings, we conclude that both adversarial loss functions and document selection strategies are more beneficial for weaker models. For smaller models, these techniques significantly improve robustness by mitigating the impact of noisy documents. For stronger models with inherently robust performance, the advantages of complex loss designs and sophisticated document selection diminish, suggesting that simpler strategies like random document selection may be sufficient. This underscores the importance of tailoring training strategies to the model’s inherent capabilities.


\subsection{Scaling Behavior and Cross-Domain Generalization}
In this section, we conduct experiments across a wide range of model scales and diverse domains to investigate how the effectiveness of robust RAG training strategies varies with model capabilities.
\subsubsection{Parameter Scale Analysis}
Through analyzing Tables~\ref{tab:main_result_llama} and~\ref{tab:main_result_qwen}, we observe a counter-intuitive phenomenon: models trained with random documents outperformed those with golden documents, despite the latter containing ground truth answers typically yielding optimal results in standard SFT. To investigate whether this phenomenon extends beyond 7B-8B models, we conduct experiments across model scales from 0.5B to 70B parameters. 

The results  in Figure~\ref{fig:parameter_scale} demonstrate that for smaller models ($\le$3B parameters), training with golden documents leads to superior performance. This suggests that smaller models, limited by their inherent capabilities, benefit more from high-quality golden documents containing direct answers. However, as model size increases, we observe that training with random documents becomes more effective. This shift can be attributed to larger models' enhanced question-answering abilities and improved robustness. These models can better generalize to downstream tasks even when trained on random documents, which may contain noisier or less structured information. This finding indicates that sophisticated document selection strategies become less crucial as model size increases, revealing an important scaling property in model training.

\subsubsection{Cross-Domain Generalization}
To validate our findings beyond general question-answering tasks, we examine whether this phenomenon generalizes across temporal reasoning (TimeQA~\cite{DBLP:conf/nips/ChenWWW21}), specialized legal knowledge (LegalBench~\cite{DBLP:conf/nips/GuhaNHRCKCPWRZT23}), and scenarios with conflicting evidence (RAGuard~\cite{worse2023}).

\begin{table}[t]
\centering
\resizebox{\linewidth}{!}{
\begin{tabular}{clcc}
\toprule
\multirow{2}{*}{\textbf{Model}} & \multirow{2}{*}{\textbf{Method}} & \multicolumn{2}{c}{\textbf{TimeQA}} \\
\cline{3-4}
& & \textbf{EM} & \textbf{F1} \\
\midrule
\multirow{5}{*}{\textit{Llama-2-7b-chat-hf}} & Golden Doc & 32.93 & 43.44 \\
& Top-1 Doc & 32.26 & 43.13 \\
& Random Doc & 31.90 & 43.32 \\
& RetRobust & 31.97 & 42.58 \\
\cmidrule{2-4}
& \textcolor{mycolor}{$\Delta$ (Worst $\rightarrow$ Best)} & \textcolor{mycolor}{3.23\%} & \textcolor{mycolor}{2.02\%} \\
\midrule
\multirow{5}{*}{\textit{Llama-3-8B-Instruct}} & Golden Doc & 35.44 & 46.90 \\
& Top-1 Doc & 35.27 & 46.94 \\
& Random Doc & 35.87 & 47.20 \\
& RetRobust & 35.90 & 47.15 \\
\cmidrule{2-4}
& \textcolor{mycolor}{$\Delta$ (Worst $\rightarrow$ Best)} & \textcolor{mycolor}{1.79\%} & \textcolor{mycolor}{0.63\%} \\
\bottomrule
\end{tabular}
}
\caption{Performance comparison of different LLMs (Llama-2-7b-chat-hf and Llama-3-8B-Instruct) across various robust RAG methods on the TimeQA dataset.}
\label{tab:timeqa_results}
\vspace{-0.5cm}
\end{table}

\paragraph{Temporal Reasoning Domain}
We evaluate the performance of various document selection strategies on the TimeQA dataset~\cite{DBLP:conf/nips/ChenWWW21}, which specifically tests models' ability to handle time-sensitive information.
As shown in Table~\ref{tab:timeqa_results}, we compare two different language models: Llama-2-7b-chat-hf and Llama-3-8B-Instruct.
With Llama-2-7b-chat-hf, we observe moderate variations across strategies, with a delta of 3.23\% in EM and 2.02\% in F1 between best and worst approaches. Golden Document selection generally yields the strongest performance. However, with Llama-3-8B-Instruct, this marginal robustness benefit narrows significantly to just 1.79\% for EM and 0.63\% for F1, with RetRobust and Random Document selection slightly outperforming Golden Document selection.

\begin{table}[t]
\centering
\resizebox{\linewidth}{!}{
\begin{tabular}{llc}
\toprule
\textbf{Model} & \textbf{Method} & \textbf{LegalBench} \\
\midrule
\multirow{9}{*}{\textit{Llama-2-7b-chat-hf}} & Base Model & 2.55 \\
& RALM & 70.96 \\
& RetRobust & 66.67 \\
& Top-1 Doc & 71.21 \\
& Golden Doc & 70.96 \\
& Random Doc & 70.96 \\
& Irrelevant Doc & 72.73 \\
& IRM & 69.19 \\
\cmidrule{2-3}
& \textcolor{mycolor}{$\Delta$ (Worst $\rightarrow$ Best)} & \textcolor{mycolor}{9.09\%} \\
\midrule
\multirow{9}{*}{\textit{Llama-3-8B-Instruct}} & Base Model & 73.23 \\
& RALM & 85.61 \\
& RetRobust & 84.85 \\
& Top-1 Doc & 85.86 \\
& Golden Doc & 86.62 \\
& Random Doc & 86.87 \\
& Irrelevant Doc & 86.36 \\
& IRM & 83.84 \\
\cmidrule{2-3}
& \textcolor{mycolor}{$\Delta$ (Worst $\rightarrow$ Best)} & \textcolor{mycolor}{3.61\%} \\
\bottomrule
\end{tabular}
}
\caption{Performance comparison of the Llama-2-7b-chat-hf and Llama-3-8B-Instruct models across various robust RAG methods on the Legal-Bench Consumer Contracts QA dataset.}
\label{tab:legal_results}
\vspace{-0.3cm}
\end{table}

\paragraph{Specialized Legal Domain}
We extend our investigation to specialized domains by evaluating various robust RAG methods on the LegalBench Consumer Contracts QA dataset~\cite{DBLP:conf/nips/GuhaNHRCKCPWRZT23}. Table~\ref{tab:legal_results} presents a comprehensive comparison across both Llama-2-7b-chat-hf and Llama-3-8B-Instruct models.
The results strongly reinforce our core findings. For Llama-2-7b-chat-hf, we observe substantial performance variations across different methods, with a 9.09\% delta between the best (Irrelevant Doc at 72.73\%) and worst (RetRobust at 66.67\%) effective approaches. This significant variation suggests that older models remain highly sensitive to the specific training methodology employed.
In contrast, the more advanced Llama-3-8B-Instruct model demonstrates higher baseline performance (73.23\% without retrieval augmentation) and reduced sensitivity to the choice of method. The performance delta between the best (Random Doc at 86.87\%) and worst effective method (IRM at 83.84\%) narrows to just 3.61\% - less than half the gap observed with Llama-2.

\begin{table}[t]
\centering
\resizebox{\linewidth}{!}{
\begin{tabular}{llc}
\toprule
\textbf{Model} & \textbf{Method} & \textbf{RAGuard} \\
\midrule
\multirow{7}{*}{\makecell[l]{\textit{Llama-2-7b-chat-hf}}} & Random Doc & 56.04 \\
& Top-1 Doc & 56.73 \\
& Golden Doc & 54.31 \\
& Irrelevant Doc & 56.00 \\
& IRM & 51.81 \\
& RetRobust & 46.71 \\
\cmidrule{2-3}
& \textcolor{mycolor}{$\Delta$ (Worst $\rightarrow$ Best)} & \textcolor{mycolor}{21.45\%} \\
\midrule
\multirow{7}{*}{\makecell[l]{\textit{Llama-3-8B-Instruct}}} & Random Doc & 59.82 \\
& Top-1 Doc & 60.08 \\
& Golden Doc & 59.37 \\
& Irrelevant Doc & 59.14 \\
& IRM & 58.95 \\
& RetRobust & 59.78 \\
\cmidrule{2-3}
& \textcolor{mycolor}{$\Delta$ (Worst $\rightarrow$ Best)} & \textcolor{mycolor}{1.92\%} \\
\bottomrule
\end{tabular}
}
\caption{Performance comparison of different LLMs (Llama-2-7b-chat-hf and Llama-3-8B-Instruct) across various robust RAG methods on the RAGuard Benchmark.}
\label{tab:raguard_results}
\vspace{-0.5cm}
\end{table}

\paragraph{Evaluation with Conflicting Evidence}
To validate our findings in challenging scenarios, we evaluate robust RAG methods on the RAGuard benchmark~\cite{worse2023}, which tests model robustness against contradictory retrievals. Table~\ref{tab:raguard_results} shows that for Llama-2-7b-chat-hf, there is a substantial 21.45\% performance gap between the best and worst training methods. In contrast, Llama-3-8B-Instruct demonstrates remarkably consistent performance, with the gap shrinking to merely 1.92\% - a 90\% reduction compared to the less capable model. These results on a benchmark with conflicting evidence further confirm our central finding that the benefits of complex robust training methods diminish significantly as model capacity increases.

Across all three domains, we consistently observe that performance variations between different document selection strategies decreases significantly with more advanced models.  These experiments confirm that as language models increase in capability, the benefits of complex robust RAG training strategies diminish significantly, regardless of domain specificity.

\section{Understanding Diminishing Benefits of Complex Training in Powerful LLMs}
In this section, we conduct comprehensive experiments to delve into the reasons why sophisticated robust training strategies may no longer be crucial in powerful models.

\begin{figure}[t]
    \centering
    \includegraphics[width=\columnwidth]{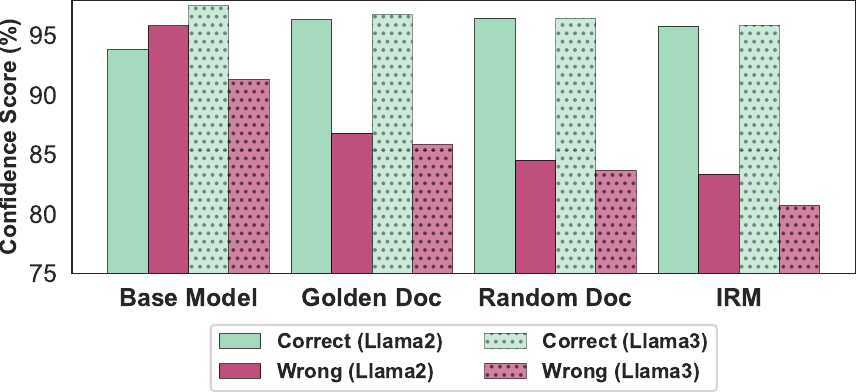}
    \caption{Confidence scores for correct and wrong answers on HotpotQA dataset, comparing Llama2 and Llama3 models across various robust training methods.}
    \label{fig:confidence}
\end{figure}

\subsection{Powerful Models Enable Natural Calibration}
To understand whether powerful models inherently possess the ability to distinguish reliable from unreliable answers, we take HotpotQA dataset as an example to examine the confidence calibration capabilities for Llama2 (Llama-2-7b-chat-hf) and Llama3 (Llama-3-8B-Instruct) models. Here, confidence is the mean of token-wise probabilities in the model's generated answer, providing a measure of the model's certainty in its predictions~\cite{DBLP:journals/corr/abs-2307-03987, DBLP:conf/iclr/XiongHLLFHH24}. Figure~\ref{fig:confidence} reveals striking differences in their calibration patterns. In the base model, Llama2 shows poor natural calibration levels, where confidence scores for incorrect answers (95.8) abnormally exceed those for correct ones (93.8). In contrast, Llama3 demonstrates inherently better calibration, maintaining higher confidence for correct answers (97.5) than incorrect ones (91.3) without any specialized complex training.

While robust training methods can effectively calibrate confidence scores and improve the gap between correct and incorrect answers for Llama2 from -2 to 12 with IRM, the marginal benefits of these complex training strategies diminish as model architectures advance.
For Llama3, which already achieves a 6.2 confidence gap naturally, the improvements from these training methods become less significant. This finding strongly suggests that advances in model architecture can effectively eliminate the need for complex robustness training procedures, as newer models come with better built-in calibration capabilities.

\begin{figure}[t]
    \centering
    \begin{subfigure}[b]{0.45\textwidth}
        \centering
        \includegraphics[width=\textwidth]{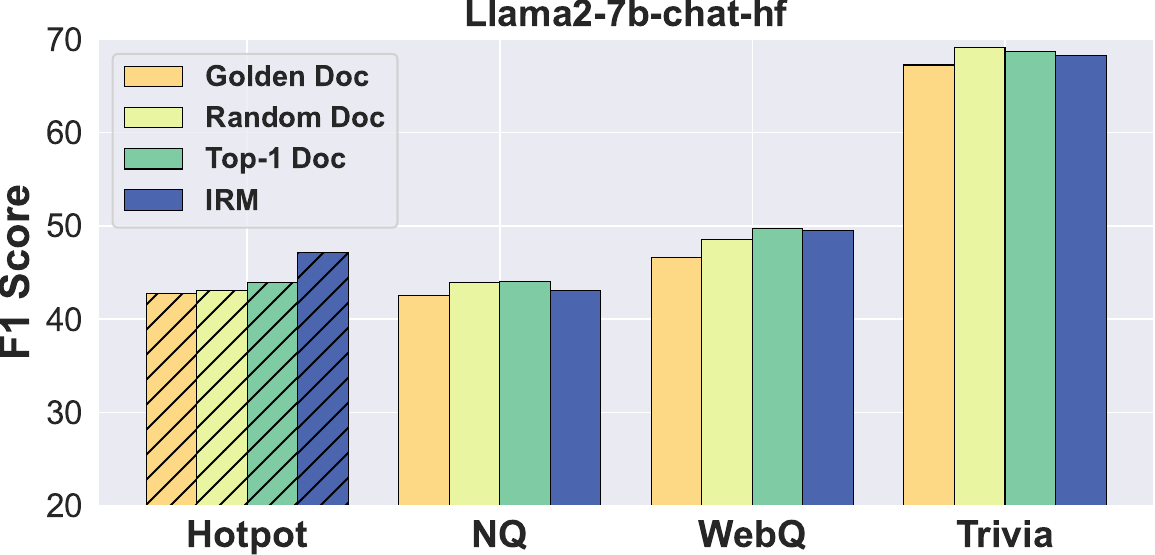}
    \end{subfigure}
    \begin{subfigure}[b]{0.45\textwidth}
        \centering
        \includegraphics[width=\textwidth]{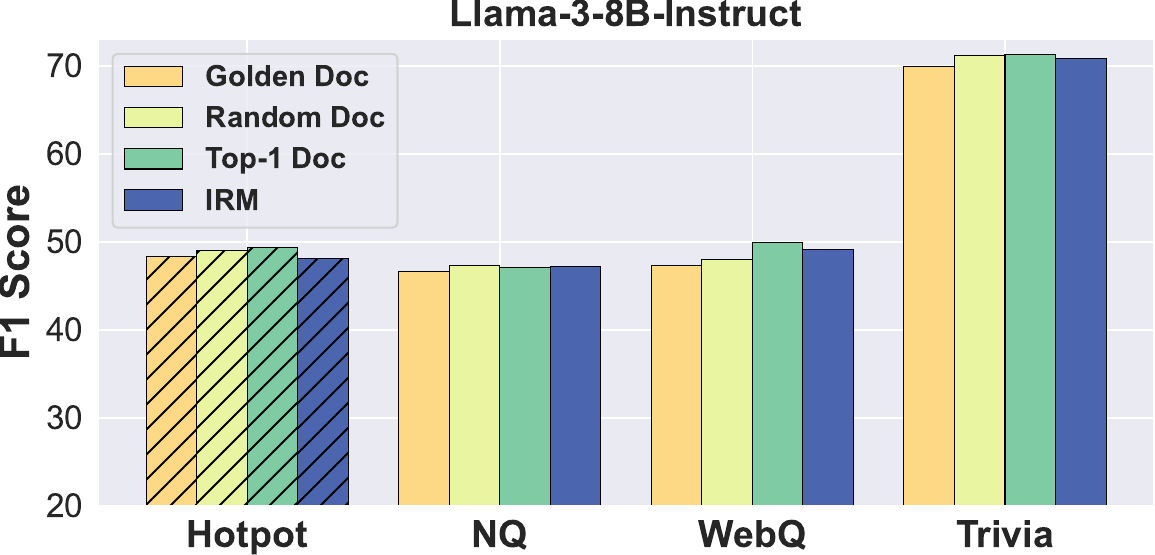}
    \end{subfigure}
    \caption{Generalization performance comparison across different strategies trained on HotpotQA (diagonal hatches bars) and evaluated on NQ, WebQuestions, TriviaQA datasets (plain bars).}
    \label{fig:cross_dataset_eval_part}
    \vspace{-0.5cm}
\end{figure}

\subsection{Simple Training Strategies Generalize Well in Powerful Models}
We further investigate whether powerful models can maintain robust generalization across different datasets with simple training strategies. We fine-tune models on HotpotQA using four document selection approaches and evaluate their transfer performance on NQ, WebQuestions, and TriviaQA.

\begin{figure}[t]
    \centering
    \includegraphics[width=\columnwidth]{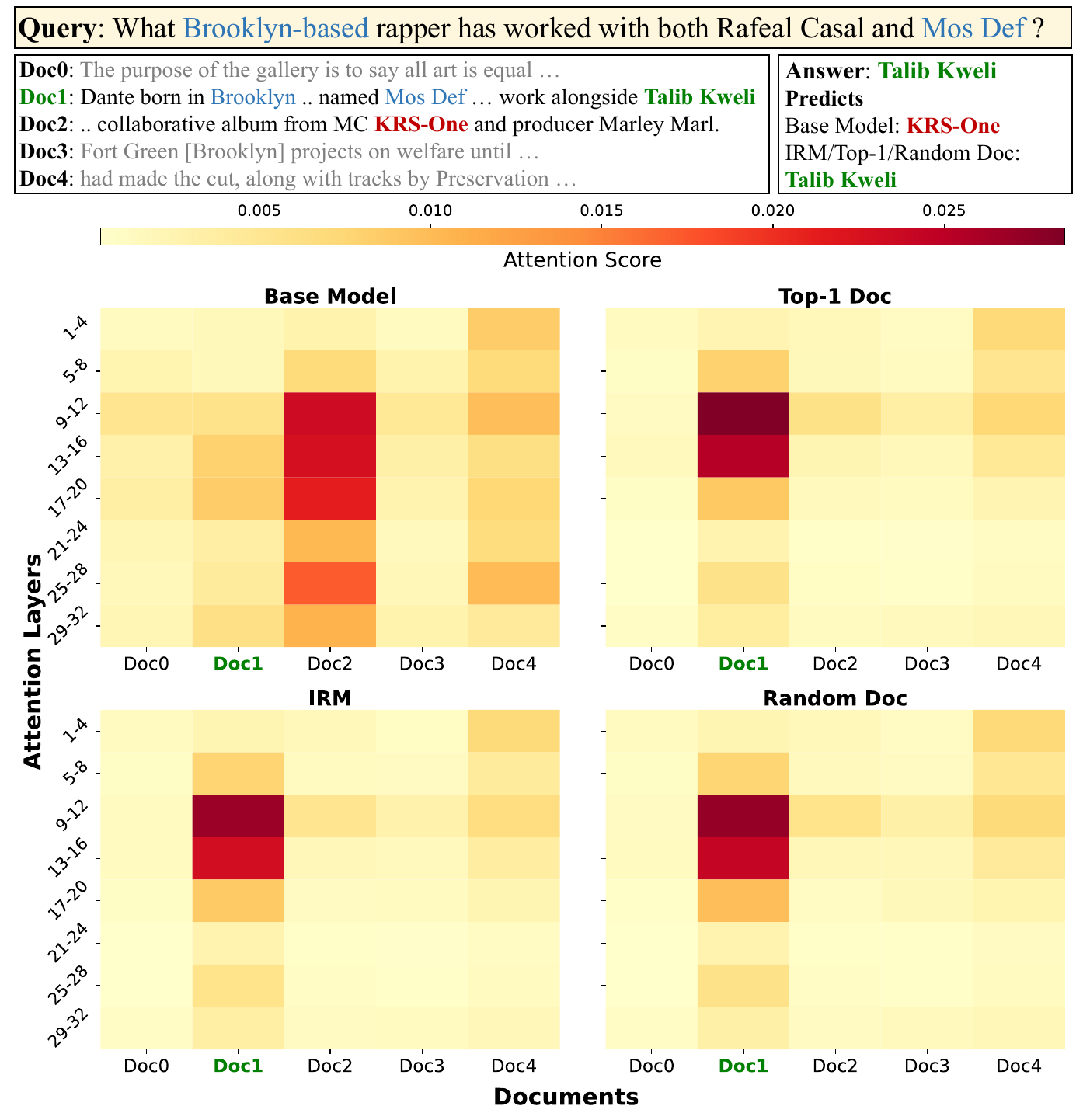}
    \caption{Attention visualization for a QA case. Each subplot shows attention distribution heatmaps across different models, where cell ($i$, $j$) represents the average attention weight from the $j$-th attention layer to document $i$. Text highlighted in \textcolor{mygreen}{green} indicates the correct answer and corresponding \textcolor{mygreen}{\textbf{Doc1}}, {blue} indicates key terms from the query, and {red} indicates incorrect model predictions. The color intensity in the heatmaps indicates attention strength.}
    \label{fig:attention}
    \vspace{-0.2cm}
\end{figure}

As shown in Figure~\ref{fig:cross_dataset_eval_part}, simple strategies demonstrate surprisingly strong generalization ability. Random document selection matches or even outperforms sophisticated IRM across all evaluation datasets, with marginal robustness benefit of less than 1\%.
For instance, in TriviaQA, random selection (69.5 F1) slightly surpasses both golden (68.2 F1) and IRM (68.7 F1) approaches. This trend becomes more pronounced in the powerful Llama-3-8B-Instruct, where the performance gap between simple and sophisticated strategies further narrows.
The consistent cross-dataset performance, regardless of training strategy, indicates that model capacity, rather than training sophistication, is the key driver of generalization ability. These findings provide strong evidence that as models become more powerful, sophisticated training strategies become increasingly unnecessary.

\subsection{Powerful Models Learn Effective Attention Patterns with Simple Training}
To provide a direct understanding of why simple training can achieve good performance, we visualize attention distributions across different training strategies. Figure~\ref{fig:attention} reveals that both sophisticated robust training methods (IRM) and simple approaches (random doc, top-1) achieve similar attention patterns, with clear focus on Doc1 (containing the correct answer) in middle layers (9-16). In contrast, the base model fails to attend to the correct document, generating a wrong answer. This finding provides direct evidence that powerful models can learn optimal attention mechanisms even with simple training strategies, making sophisticated training methods unnecessary.

\begin{figure}[t]
    \centering
    \includegraphics[width=\columnwidth]{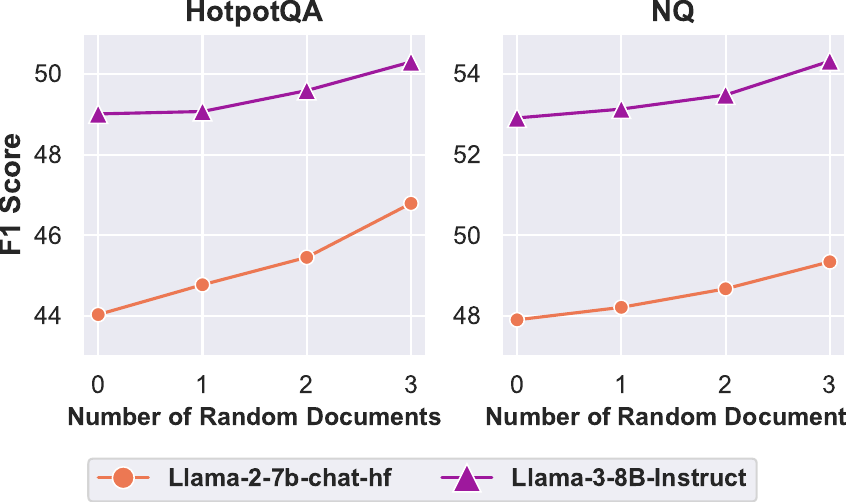}
    \caption{Performance comparison with different numbers of random documents during training. Increasing the number of random documents consistently improves model performance.}
    \label{fig:mixed_doc}
    \vspace{-0.3cm}
\end{figure}

\begin{figure}[t]
    \centering
    \includegraphics[width=\columnwidth]{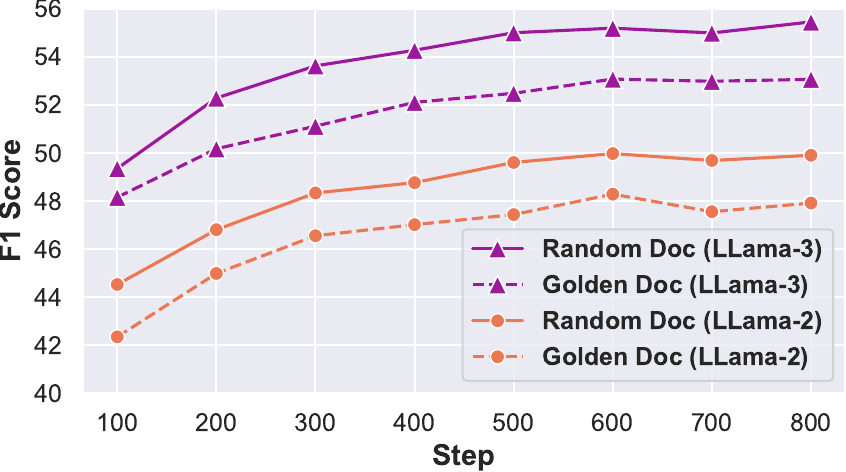}
    \caption{Training curves comparison between random and golden document strategies using Llama-2-7b-chat-hf and LLama-3-8B-Instruct.}
    \label{fig:training_step_f1}
    \vspace{-0.3cm}
\end{figure}

\subsection{Training with Random Docs: Better Performance and Faster Convergence}
We investigate why random training proves good performance from two aspects.
\paragraph{More random docs lead to better performance}
We first vary the numbers of random documents (0 to 3) in training instances to examine how increasing random documents affects model performance. As shown in Figure~\ref{fig:mixed_doc}, increasing random documents consistently improves F1 scores across both datasets. For Llama-2-7b-chat-hf, using 3 random documents (versus zero) improves F1 scores by 3 points on HotpotQA and 4 points on NQ. Llama-3-8B-Instruct shows similar gains, suggesting that powerful models can effectively learn from random documents, making sophisticated document selection relatively unnecessary.

\paragraph{Faster convergence with random training}
The training dynamics in Figure~\ref{fig:training_step_f1} provide another evidence for why sophisticated document selection becomes unnecessary. Random document training not only achieves higher F1 scores (2-3 points improvement) but also reaches peak performance in fewer steps compared to golden document training. This faster convergence with better performance holds true for both model scales, indicating that simpler random training actually enables more efficient learning in powerful language models.

\section{Conclusion}
Our study reveals that as language model capacity increases, the benefits of complex robust RAG training strategies diminish significantly. We observe a counter-intuitive phenomenon: \emph{as model capability increases, the marginal robustness benefit of complex training diminishes sharply, and random document training often matches or exceeds the performance of complex methods}. This effect generalizes across temporal reasoning, legal knowledge, and scenarios with conflicting evidence. Our analysis shows that stronger models possess inherent capabilities—better confidence calibration, cross-dataset generalization, and effective attention patterns—that reduce their reliance on complex training. These findings enable simplified RAG system design through streamlined retrieval mechanisms and open the door to more scalable systems that can effectively learn from noisy retrieval results without elaborate document filtering.

\section*{Acknowledgement}
This work was supported by the Strategic Priority Research Program of the CAS under Grants No.XDB0680302, the National Natural Science Foundation of China (NSFC) under Grants No. 62276248, the Key Research and Development Program of Xinjiang Uyghur Autonomous Region Grant No. 2024B03026, the Beijing Nova Program under Grants No. 20250484765, and the Youth Innovation Promotion Association CAS under Grants No. 2023111.


\bibliographystyle{ACM-Reference-Format}
\balance
\bibliography{sample-base}

\end{document}